\documentclass[a4paper,fleqn]{cas-dc}
\usepackage[numbers,sort&compress]{natbib}
\usepackage{cas-common}

\usepackage{soul}

\hypersetup{colorlinks=true, linkcolor=blue, anchorcolor=blue, citecolor=blue}

\def\tsc#1{\csdef{#1}{\textsc{\lowercase{#1}}\xspace}}
\tsc{WGM}
\tsc{QE}
\tsc{EP}
\tsc{PMS}
\tsc{BEC}
\tsc{DE}

\begin{document}
\let\printorcid\relax
\let\WriteBookmarks\relax
\def\floatpagepagefraction{1}
\def\textpagefraction{.001}

\shorttitle{REASON: P\underline{R}obability Map-Guid\underline{E}d Du\underline{A}l-Branch Fu\underline{S}i\underline{ON} Framework for Gastric Content Assessment}

\shortauthors{N.-F. Xiao et~al.}

\title [mode = title]{REASON: P\underline{R}obability Map-Guid\underline{E}d Du\underline{A}l-Branch Fu\underline{S}i\underline{ON} Framework for Gastric Content Assessment}

\author[1,2]{Nu-Fang Xiao}\fnmark[$\dagger$]
\ead{23010501022@mail.hnust.edu.cn}
\credit{Data curation, Methodology, Writing - original draft}

\author[2,3]{De-Xing Huang}\fnmark[$\dagger$]
\ead{huangdexing2022@ia.ac.cn}
\credit{Data curation, Methodology, Writing - review \& editing}

\author[4]{Le-Tian Wang}\fnmark[$\dagger$]
\ead{wanglt5551@163.com}
\credit{Data curation, Methodology}

\author[2,3]{Mei-Jiang Gui}
\credit{Resources}

\author[1]{Qi Fu}
\credit{Resources}

\author[2,3]{Xiao-Liang Xie}
\credit{Resources}

\author[2,3]{Shi-Qi Liu}
\credit{Resources}

\author[2,3]{Shuangyi Wang}
\credit{Resources}

\author[2,3]{Zeng-Guang Hou}
\credit{Resources, Funding acquisition}

\author[4]{Ying-Wei Wang}\cormark[1]
\ead{wangyw@fudan.edu.cn}
\credit{Supervision}

\author[2,3]{Xiao-Hu Zhou}\cormark[1]
\ead{xiaohu.zhou@ia.ac.cn}
\credit{Supervision, Conceptualization}

\affiliation[1]{organization={School of Computer Science and Engineering, Hunan University of Science and Technology},
city={Xiangtan},
postcode={411199},
country={China}}

\affiliation[2]{organization={State Key Laboratory of Multimodal Artificial Intelligence Systems, Institute of Automation, Chinese Academy of Sciences},
city={Beijing},
postcode={100190},
country={China}}

\affiliation[3]{organization={School of Artificial Intelligence, University of Chinese Academy of Sciences},
city={Beijing},
postcode={100049},
country={China}}

\affiliation[4]{organization={Department of Anesthesiology, Huashan Hospital, Fudan University},
city={Shanghai},
postcode={200040},
country={China}}

\nonumnote{$^\dagger$These authors contributed equally to this work.}
\nonumnote{$^*$Corresponding authors.}

\begin{abstract}
Accurate assessment of gastric content from ultrasound is critical for stratifying aspiration risk at induction of general anesthesia. However, traditional methods rely on manual tracing of gastric antra and empirical formulas, which face significant limitations in both efficiency and accuracy. To address these challenges, a novel two-stage p\underline{R}obability map-guid\underline{E}d du\underline{A}l-branch fu\underline{S}i\underline{ON} framework (REASON) for gastric content assessment is proposed. In stage 1, a segmentation model generates probability maps that suppress artifacts and highlight gastric anatomy. In stage 2, a dual-branch classifier fuses information from two standard views, right lateral decubitus (RLD) and supine (SUP), to improve the discrimination of learned features. Experimental results on a self-collected dataset demonstrate that the proposed framework outperforms current state-of-the-art approaches by a significant margin. This framework shows great promise for automated preoperative aspiration risk assessment, offering a more robust, efficient, and accurate solution for clinical practice.
\end{abstract}



\maketitle

\section{Introduction}~\label{sec:introduction}
\begin{figure}[t]
\centering
\includegraphics{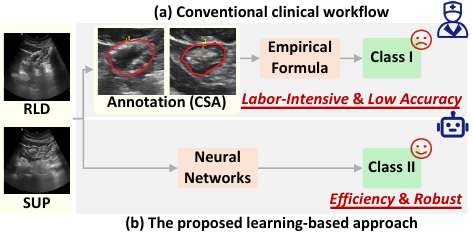}
\caption{\textbf{Comparison between the conventional clinical workflow and the proposed method.} (a) The conventional clinical workflow, which relies on empirical formulas, is labor-intensive and suffers from limited generalizability. (b) In contrast, the proposed learning-based approach offers superior efficiency and robustness.}
\label{fig:comparison}
\end{figure}

Pulmonary aspiration is a potentially life-threatening complication, particularly during general anesthesia or procedural sedation~\cite{wang2021gastric,warner2021pulmonary,sherwin2023influence}. Reliable preoperative assessment of gastric content is therefore essential for perioperative risk management~\cite{van2014ultrasound}. In clinical practice, gastric ultrasound is widely used to estimate gastric content by measuring the cross-sectional area (CSA) of the antrum~\cite{king2019preoperative,frykholm2022pre,demirel2023ultrasonographic,baettig2023pre}. The standard procedure involves: \textit{\textbf{i)}} manual delineation of the antral contour to calculate the CSA, \textit{\textbf{ii)}} estimation of gastric volume using empirical formulas, and \textit{\textbf{iii)}} classification of gastric content into predefined grades, as shown in Fig.~\ref{fig:comparison} (a). These conventional approaches have significant limitations. Accurate annotation depends on anesthesiologists' expertise, making the process time-consuming and labor-intensive~\cite{liu2020semi}. In addition, the empirical formulas used for volume estimation are sensitive to image quality and patient-specific variability, which reduces generalizability and accuracy~\cite{kruisselbrink2019diagnostic}. These challenges underscore the necessity for automated gastric content assessment methods.

Recent advances in deep learning have achieved strong performance across natural and medical image analysis~\cite{dhar2023challenges,wei2021fine,chen2022recent, huang2025real}. However, gastric ultrasound poses unique challenges for learning-based approaches, including pronounced speckle noise, acoustic artifacts, and high morphological variability of gastric antra~\cite{khan2022experimental}. Existing methods frequently \textit{\textbf{fail to localize gastric regions}} reliably and have difficulty learning robust features~\cite{van2019deep}. In addition, gastric ultrasound acquisition typically includes two complementary views—right lateral decubitus (RLD) and supine (SUP)—which provide distinct spatial and contextual information. Yet current approaches \textbf{\textit{lack effective strategies to integrate multi-view images}}, limiting the discriminative power of learned features.

\begin{figure*}[t]
\centering
\centerline{\includegraphics{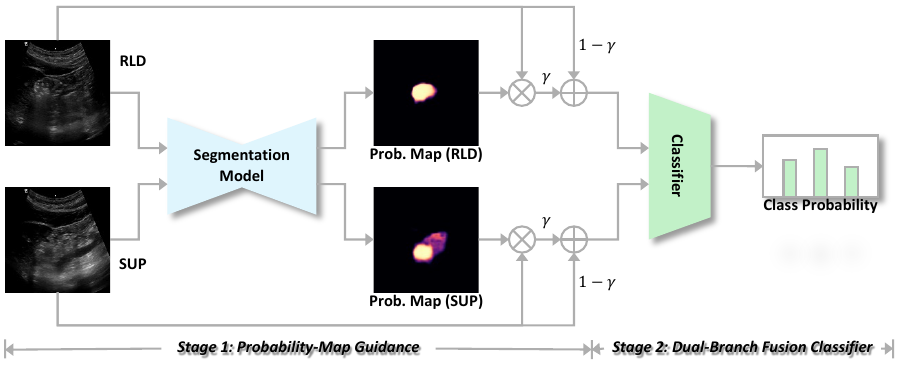}}
\caption{\textbf{Overall framework of REASON.} It consists of two stages. In stage 1, probability maps generated by the segmentation model are used to highlight anatomically relevant gastric regions and suppress artifacts. In stage 2, the dual-branch fusion classifier integrates complementary spatial and contextual features from the RLD and SUP views to enable robust gastric content assessment.}
\label{fig:reason_framework}
\end{figure*}

To address these challenges, a two-stage p\underline{R}obability map-guid\underline{E}d du\underline{A}l-branch fu\underline{S}i\underline{ON} framework (REASON) for automated gastric content assessment is proposed, as illustrated in Fig.~\ref{fig:comparison} (b). In the first stage, probability map guidance (PMG) is introduced, which leverages probability maps generated by a segmentation model to enhance the model’s focus on gastric regions. To reduce the annotation burden for training the segmentation model, a semi-supervised learning strategy based on the mean-teacher paradigm is employed. In the second stage, a dual-branch fusion classifier (DBFC) processes the SUP and RLD views in parallel, and their predictions are fused to exploit complementary information.

The contributions are summarized as follows:
\begin{itemize}
    \item A two-stage p\underline{R}obability map-guid\underline{E}d du\underline{A}l-branch fu\underline{S}i\underline{ON} framework (REASON) for gastric content assessment is proposed. To the best of our knowledge, this is the first application of deep learning techniques in this domain.
    \item A probability map-guided approach is introduced to highlight anatomically relevant regions while suppressing artifacts in gastric ultrasound. In addition, a dual-branch fusion classifier is designed to effectively integrate complementary spatial and contextual information from multiple ultrasound views.
    \item Comprehensive experiments on a self-collected dataset demonstrate REASON's superiority over existing approaches.
\end{itemize}

The remainder of this paper is organized as follows. Section~\ref{sec:related_works} reviews related studies. Section~\ref{sec:method} presents the proposed framework in detail. Section~\ref{sec:results} reports the experimental results and analysis. Finally, Section~\ref{sec:conclusion} summarizes the work and outlines future directions.

\section{Related Work}~\label{sec:related_works}
This section reviews related work in three areas: gastric content assessment (Section~\ref{sec:rw_1}), medical image segmentation (Section~\ref{sec:rw_2}), and medical image classification (Section~\ref{sec:rw_3}).

\subsection{Gastric Content Assessment}\label{sec:rw_1}
Gastric content assessment has long been an important topic in clinical practice. Existing approaches primarily rely on imaging modalities such as magnetic resonance imaging (MRI), computed tomography (CT), and gastric ultrasound. MRI and CT provide high spatial resolution~\cite{bharucha2014comparison, okada2020clinical}, enabling accurate segmentation of the three-dimensional gastric lumen for precise volume estimation. However, MRI and CT face significant barriers to routine use: MRI is time-consuming, costly, and unsuitable for point-of-care assessment, whereas CT exposes patients to ionizing radiation and is impractical in perioperative or emergency settings~\cite{sodickson2009recurrent, albert2013radiation}.

\begin{figure*}[t]
\centering
\includegraphics{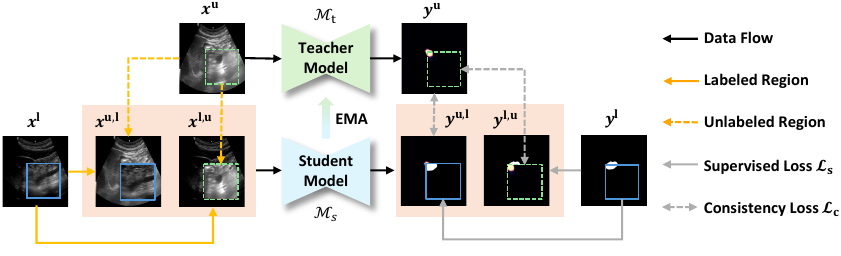} 
\caption{\textbf{Bidirectional copy-paste strategy.} BCP generates composite images that combine regions of strong supervision (from ground truth) with weak supervision (from pseudo-labels), allowing the former to effectively regularize and improve the latter.}
\label{fig:bcp} 
\end{figure*}

In contrast, ultrasound has become the preferred modality due to its non-invasiveness, safety, portability, real-time imaging, and cost-effectiveness~\cite{kruisselbrink2019diagnostic}. Kruisselbrink \textit{et al.}~\cite{kruisselbrink2014intra} demonstrated the reliability and consistency of gastric content assessment using two methods: measurement of two orthogonal diameters and freehand tracing. Perlas \textit{et al.}~\cite{perlas2016aim} further proposed an empirical formula based on the CSA of the gastric antrum for gastric content estimation. Despite these advances, conventional methods are still hindered by labor-intensive manual annotation and limited generalizability across patient populations and imaging conditions. 

\subsection{Medical Image Segmentation}~\label{sec:rw_2} Medical image segmentation has become a cornerstone of many clinical applications~\cite{qureshi2023medical}. Fully supervised approaches typically employ deep networks trained end-to-end with pixel-wise annotations~\cite{mo2022review,ronneberger2015u,chen2021transunet,huang2024spironet}. However, these advanced models still rely heavily on extensive, high-quality manual annotations, which are costly and time-consuming to obtain. Semi-supervised segmentation has therefore emerged as a promising alternative~\cite{jiao2024learning,chen2021semi,you2023rethinking}. Existing methods can be broadly divided into two categories: pseudo-labeling~\cite{yang2022survey,wang2022semi} and consistency regularization~\cite{zhang2021flexmatch,cascante2021curriculum}. In pseudo-labeling, a model trained on labeled data generates pseudo-labels for unlabeled images, which are then used to augment the training set. Since model performance is highly dependent on pseudo-label accuracy, improving the reliability of pseudo-labels remains a central focus~\cite{han2024deep,xu2022bayesian,yao2022enhancing}. Instead, consistency regularization enforces prediction stability under different perturbations of the same image, often within the mean-teacher (MT) framework~\cite{tarvainen2017mean,xu2023ambiguity,zhang2023multi}. In this study, probability maps from segmentation models are leveraged to guide attention toward gastric regions, and an MT-based semi-supervised strategy is adopted to reduce annotation requirements while maintaining robust performance.

\subsection{Medical Image Classification}~\label{sec:rw_3}
Medical image classification aims to assign clinically meaningful labels to input images~\cite{zhang2019medical,miranda2016survey}. Early approaches primarily adapted models originally developed for natural image classification, leveraging general-purpose architectures~\cite{he2016deep, huang2017densely, tan2019efficientnet, liu2023soft} pretrained on large-scale datasets such as ImageNet~\cite{deng2009imagenet} and transferred to medical images via transfer learning. However, the performance of these approaches was constrained by the substantial domain gap between natural and medical images~\cite{dufumier2021contrastive}. To address this issue, subsequent work increasingly focused on the design of domain-specific architectures tailored to the unique characteristics of medical imaging~\cite{yue2024medmamba}. Advances include enhanced feature representation learning~\cite{wu2023exploring}, specialized regularization strategies~\cite{manzari2023medvit}, and innovative training paradigms~\cite{gao2024training}, which collectively enable models to better capture the intrinsic properties of medical data. Despite these advances, existing methods remain inadequate for gastric content assessment. Specifically, current models struggle to extract reliable gastric features from noisy ultrasound and do not effectively integrate complementary information from multi-view inputs. To mitigate these gaps, PMG and DBFC are introduced: PMG suppresses noise and highlights gastric-relevant regions, whereas DBFC integrates complementary information from multi-view inputs.

\section{Method}~\label{sec:method}
The overall framework of REASON is illustrated in Fig.~\ref{fig:reason_framework}. It consists of two stages: probability map guidance (PMG) and the dual-branch fusion classifier (DBFC). In the first stage (Section~\ref{sec:pmg}), a segmentation model trained under the mean-teacher framework generates pixel-level probability maps $p \in [0,1]^{C_0 \times H \times W}$, where $C_0$, $H$, and $W$ denote the number of channels, height, and width, respectively. These probability maps are then used to guide the RLD image $x_{\rm r}\in\mathbb{R}^{C\times H\times W}$ and the SUP image $x_{\rm s}\in\mathbb{R}^{C\times H\times W}$, enhancing the gastric regions. In the second stage (Section~\ref{sec:dbfc}), the DBFC integrates the enhanced multi-view representations to achieve accurate and robust classification.

\subsection{Stage 1: Probability Map Guidance (PMG)}~\label{sec:pmg}
Because gastric ultrasound often exhibits a low signal-to-noise ratio (SNR), existing approaches struggle to learn robust representations for accurate content assessment. In the first stage, probability maps generated by a segmentation model are leveraged to highlight gastric regions and suppress noise.

To reduce the annotation burden, a semi-supervised strategy based on the mean-teacher (MT)~\cite{tarvainen2017mean} framework is adopted to train the segmentation model. The training dataset is defined as $\mathcal{D}=\mathcal{D}^{\rm l}\cup \mathcal{D}^{\rm u}$, where $\mathcal{D}^{\rm l}=\left\{x^{\rm l}_i, y^{\rm l}_i\right\}_{i=1}^M$ is the labeled subset and $\mathcal{D}^{\rm u}=\left\{x^{\rm u}_i\right\}_{i=M+1}^N$ is the unlabeled subset, with $M\ll N$. Here, $y^{\rm l}_i \in \{0,1\}^{H\times W}$ denotes the pixel-level annotation mask.

First, a U-Net~\cite{ronneberger2015u} is pretrained on the labeled subset and used as the backbone for both the teacher and student networks. Following the vanilla MT framework~\cite{tarvainen2017mean}, the objective of the semi-supervised task can be formulated as:
\begin{align}
    \min_{\theta_s,\theta_t} \Big[
        \mathcal{L}_{\rm s}(\theta_s, \mathcal{D}^{\rm l}) 
        + \mathcal{L}_{\rm c}(\theta_s, \theta_t, \mathcal{D}^{\rm u})
    \Big]
\end{align}
where $\theta_\mathrm{s}$ and $\theta_\mathrm{t}$ denote the parameters of the student and teacher networks, respectively, $\mathcal{L}_{\rm s}$ denotes the supervised loss on $\mathcal{D}^{\rm l}$, and $\mathcal{L}_{\rm c}$ represents the consistency loss on $\mathcal{D}^{\rm u}$.

To improve the robustness of the MT framework, we employ the Bidirectional Copy-Paste (BCP) strategy~\cite{bai2023bidirectional}. As illustrated in Fig.~\ref{fig:bcp}, BCP operates by randomly selecting a labeled image $x^{\rm l}$ and an unlabeled image $x^{\rm u}$ from the training dataset. A random patch, covering approximately 25\% of the total area, is cropped from each image. These patches are then swapped to create two new composite images, $x^{\rm u,l}$ and $x^{\rm l,u}$:
\begin{align}
    x^{\rm u,l} &= x^{\rm u} \odot m + x^{\rm l} \odot (1 - m) \\
    x^{\rm l,u} &= x^{\rm l} \odot m + x^{\rm u} \odot (1 - m)
\end{align}
where $m \in \{0,1\}^{H\times W}$ is a binary mask defining the patch region and $\odot$ denotes element-wise multiplication.

Both composite images are then fed into the student model $\mathcal{M}_{\rm s}$ to generate segmentation predictions, $y^{\rm u, l}$ and $y^{\rm l, u}$, respectively:
\begin{align}
    y^{\rm u, l} = \mathcal{M}_{\rm s}\left(x^{\rm u, l}\right), y^{\rm l, u} = \mathcal{M}_{\rm s}\left(x^{\rm l, u}\right)
\end{align}

The total training objective $\mathcal{L}_{\rm seg}$ is a composite loss that enforces supervised learning on regions originating from the labeled image and consistency regularization on regions from the unlabeled image:
\begin{equation}
    \mathcal{L}_{\rm seg} = \mathcal{L}_{\rm s} + \mathcal{L}_{\rm c}
\end{equation}

Specifically, the supervised loss $\mathcal{L}_{\rm s}$ is calculated on the labeled-origin pixels within both composite images, using the ground-truth mask $y^{l}$:
\begin{equation}
\mathcal{L}_{\rm s} = \mathcal{L}\left[y^{\rm u,l} \odot \left(1-m\right), y^{\rm l} \odot \left(1-m\right)\right] + \mathcal{L}\left[y^{\rm l,u} \odot m, y^{\rm l} \odot m\right]
\end{equation}

Concurrently, the consistency loss $\mathcal{L}_{\rm c}$ is applied to the unlabeled-origin regions. It enforces that the student's predictions for these areas are consistent with the pseudo-labels $y^{\rm u}$, which are generated by the teacher model $\mathcal{M}_{\rm t}$:
\begin{equation}
\mathcal{L}_{\rm c} = \mathcal{L}\left[y^{\rm u,l} \odot m, y^{\rm u} \odot m\right] + \mathcal{L}\left[y^{\rm l,u} \odot (1-m), y^{\rm u} \odot (1-m)\right]
\end{equation}

The base loss function $\mathcal{L}$ for both terms is a combination of the Dice loss $\mathcal{L}_{\rm Dice}$ and the cross-entropy loss $\mathcal{L}_{\rm ce}$:
\begin{equation}
\mathcal{L} = \mathcal{L}_{\rm Dice} + \mathcal{L}_{\rm ce}
\end{equation}

This process is integrated into the MT framework. While the student model's parameters $\theta_{\rm s}$ are updated via backpropagation, the teacher model's parameters $\theta_{\rm t}$ are not trained directly. Instead, they are updated using an Exponential Moving Average (EMA) of the student's parameters at each iteration $k$:
\begin{equation}
\theta_\mathrm{t}^{(k+1)} = \alpha\, \theta_\mathrm{t}^{(k)} + (1 - \alpha)\, \theta_\mathrm{s}^{(k)}
\end{equation}
where the hyper-parameter $\alpha$ is set to 0.99 following previous work~\cite{bai2023bidirectional}.

\begin{figure}[t]
\centering
\includegraphics{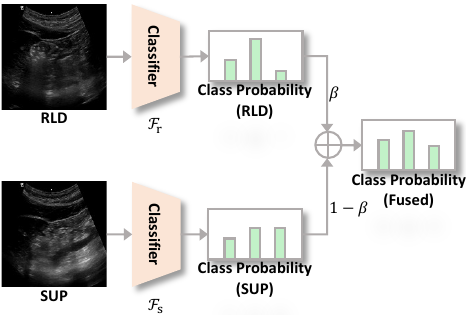}
\caption{\textbf{Dual-branch fusion classifier.} It integrates complementary spatial and contextual information from SUP and RLD views through a weighted combination of logits.
}
\label{fig:dbf}
\end{figure}

\subsection{Stage 2: Dual-Branch Fusion Classifier (DBFC)}~\label{sec:dbfc}
In the second stage, a dual-branch fusion classifier is introduced to integrate multi-view ultrasound images, as illustrated in Fig.~\ref{fig:dbf}. Each input image is first enhanced using probability maps generated in \textit{\textbf{Stage 1}}, which highlight the gastric regions.
\begin{align}
    x^{\prime}=(1-\gamma)\cdot x+\gamma\cdot x\odot p
\end{align}
where $\gamma\in [0,1]$ is a hyper-parameter that controls the strength of the attention mechanism. In implementation, $\gamma$ is set to 0.5 based on the ablation study results.

The multi-view enhanced images $x^{\prime}_{\rm r}$ and $x^{\prime}_{\rm s}$ are then fed into two sub-classifiers, $\mathcal{F}_{\rm r}$ and $\mathcal{F}_{\rm s}$, to obtain class-wise probabilities:
\begin{align}
    y_{\rm r}={\rm Softmax}\left[\mathcal{F}_{\rm r}\left(x^{\prime}_{\rm r}\right)\right], y_{\rm s}={\rm Softmax}\left[\mathcal{F}_{\rm s}\left(x^{\prime}_{\rm s}\right)\right]
\end{align}
where $y_{\rm r}, y_{\rm s} \in [0,1]^{n_{\rm cls}}$ and $n_{\rm cls}$ denotes the number of classes. Each sub-classifier uses DenseNet121~\cite{huang2017densely} as the backbone.

To integrate information from the two imaging views, the class-wise probabilities from the two branches are fused using a weighted sum:
\begin{align}
y_{\rm f} = \beta \cdot y_{\rm r} + (1-\beta) \cdot y_{\rm s},
\end{align}
where $\beta \in [0,1]$ is a hyper-parameter that controls the relative importance of each view. Based on extensive ablation studies, $\beta$ is set to 0.7.

Because the dataset is class-imbalanced, focal loss~\cite{lin2017focal} is adopted to train the classifier. The overall training objective combines the fusion loss with the individual branch losses:
\begin{equation}
\mathcal{L}_{\rm cls} = \mathcal{L}_{\rm Focal}(y_{\rm f}, y_{\rm gt}) + u\left[\mathcal{L}_{\rm Focal}(y_{\rm r}, y_{\rm gt}) + \mathcal{L}_{\rm Focal}(y_{\rm s}, y_{\rm gt})\right]
\end{equation}
where $u$ is a hyper-parameter that balances the contributions of different supervision signals, and it is set to 0.3 by default.

\section{Results}~\label{sec:results}
\subsection{Dataset}~\label{sec:criteria}

\begin{table*}[t]
\caption{\textbf{Comparison of the state-of-the-art methods on the ultrasound dataset.} The best results are highlighted in \textbf{bold} and the second best results are \underline{underlined}. Results of all baseline methods are averaged over two image views. $^\dagger$ indicates that the model is specialized in medical imaging. $^{**}$ indicates statistical significance at $p < 0.01$.}
\label{table:sota}
\centering
\renewcommand\arraystretch{1.2}
\begin{tabular}{lllll}
\toprule
Method & Acc. (\%) $\uparrow$ & Pre. (\%) $\uparrow$ & Rec. (\%) $\uparrow$ & F1 (\%) $\uparrow$ \\ \midrule
Empirical Formula \cite{van2014ultrasound} {\tiny \color{gray} [BJA'14]} & 52.75 & 52.48 & 52.62 & 52.53 \\
VGG16 \cite{simonyan2014very} {\tiny \color{gray} [ICLR'15]} & 62.26{\tiny$\pm$24.87} & 56.41{\tiny$\pm$35.26} & 62.47{\tiny$\pm$25.77} & 56.65{\tiny$\pm$32.49} \\
ResNet50 \cite{he2016deep} {\tiny \color{gray} [CVPR'16]} & 68.42{\tiny$\pm$4.60} & 70.07{\tiny$\pm$4.83} & 69.18{\tiny$\pm$4.79} & 68.33{\tiny$\pm$4.19} \\
ResNet101 \cite{he2016deep} {\tiny \color{gray} [CVPR'16]} & 69.04{\tiny$\pm$4.16} & 70.84{\tiny$\pm$5.25} & 69.43{\tiny$\pm$4.28} & 68.47{\tiny$\pm$3.48} \\
DenseNet121 \cite{huang2017densely} {\tiny \color{gray} [CVPR'17]} & 71.63{\tiny$\pm$2.86} & 73.03{\tiny$\pm$4.26} & 72.15{\tiny$\pm$3.30} & 71.35{\tiny$\pm$2.41} \\
DenseNet169 \cite{huang2017densely} {\tiny \color{gray} [CVPR'17]} & 70.93{\tiny$\pm$3.55} & 72.36{\tiny$\pm$4.26} & 71.70{\tiny$\pm$3.64} & 70.84{\tiny$\pm$3.12} \\
EfficientNet-B0 \cite{tan2019efficientnet} {\tiny \color{gray} [ICML'19]} & 72.92{\tiny$\pm$5.00} & 74.21{\tiny$\pm$5.94} & 73.68{\tiny$\pm$5.15} & 72.87{\tiny$\pm$5.06} \\
EfficientNet-B5 \cite{tan2019efficientnet} {\tiny \color{gray} [ICML'19]} & 63.47{\tiny$\pm$6.90} & 66.13{\tiny$\pm$6.03} & 64.22{\tiny$\pm$6.94} & 63.63{\tiny$\pm$6.86} \\
MobileNet-V3 \cite{howard2019searching} {\tiny \color{gray} [ICCV'19]} & 71.70{\tiny$\pm$3.00} & 73.04{\tiny$\pm$3.70} & 72.26{\tiny$\pm$3.56} & 71.64{\tiny$\pm$2.73} \\
ViT-B \cite{dosovitskiy2020image} {\tiny \color{gray} [ICLR'21]} & 35.74{\tiny$\pm$3.45} & 12.91{\tiny$\pm$2.67} & 33.30{\tiny$\pm$0.24} & 18.34{\tiny$\pm$2.46} \\
Swin-B \cite{liu2021swin} {\tiny \color{gray} [ICCV'21]} & 36.09{\tiny$\pm$3.62} & 14.00{\tiny$\pm$4.8} & 33.53{\tiny$\pm$0.4} & 18.09{\tiny$\pm$1.98} \\
ConvNeXt-B \cite{liu2022convnet} {\tiny \color{gray} [CVPR'22]} & 53.51{\tiny$\pm$10.18} & 54.14{\tiny$\pm$9.88} & 53.29{\tiny$\pm$10.80} & 52.32{\tiny$\pm$10.39} \\
SoftAug \cite{liu2023soft} {\tiny \color{gray} [CVPR'23]} & \underline{74.93}{\tiny$\pm$3.47}$^{\ast\ast}$ & \underline{76.07}{\tiny$\pm$3.52}$^{\ast\ast}$ & 75.35{\tiny$\pm$3.78} & \underline{74.89}{\tiny$\pm$3.73}$^{\ast\ast}$ \\
MedMamba$^\dagger$ \cite{yue2024medmamba} {\tiny \color{gray} [arXiv'24]} & 74.51{\tiny$\pm$6.77} & 75.98{\tiny$\pm$5.90} & \underline{75.47}{\tiny$\pm$6.49} & 74.82{\tiny$\pm$6.35} \\
HiFuse$^\dagger$ \cite{huo2024hifuse} {\tiny \color{gray} [BSPC'24]} & 63.16{\tiny$\pm$7.60} & 64.40{\tiny$\pm$17.20} & 62.38{\tiny$\pm$8.10} & 57.02{\tiny$\pm$11.69} \\
REASON {\tiny \color{gray} \textbf{[Ours]}} & \textbf{82.15}{\tiny$\pm$3.98} & \textbf{83.23}{\tiny$\pm$2.89} & \textbf{82.82}{\tiny$\pm$3.35} & \textbf{82.10}{\tiny$\pm$3.98} \\
\bottomrule
\end{tabular}
\end{table*}

The gastric ultrasound dataset is collected at Huashan Hospital, Fudan University, and comprises 2,174 images from 364 patients, including 1,087 SUP and 1,087 RLD views. The dataset is split into training, validation, and test sets in a 7:2:1 ratio at the patient level.

\begin{table}[htbp]
\caption{
\textbf{Statistics of the gastric content assessment task.}
}
\label{table:data_statistics}
\centering
\renewcommand\arraystretch{1.2}
\renewcommand\arraystretch{1.2}
\begin{tabular}{ll}
\toprule
Class & Num. \\ \midrule
I ($V\leq$ 50 ml) & 868 \\
II (50 ml$<V\leq$ 100 ml) & 664 \\
III ($V>$ 100 ml) & 642 \\ \bottomrule
\end{tabular}
\end{table}

\textbf{Ethics Approval.} The study protocol receives approval from the Ethics Committee of Huashan Hospital, Fudan University. Written informed consent is obtained from all participants prior to data collection. All images are de-identified during preprocessing, and strict anonymization protocols are applied to protect patient privacy and confidentiality. All procedures adhere to the principles of the Declaration of Helsinki.

\textbf{Gastric Annotation.} All images are manually annotated by an experienced anesthesiologist using Labelme~\cite{russell2008labelme}, yielding pixel-level segmentation masks of the gastric regions. Within the semi-supervised setting, only 10\% of the labeled images from the training set are used, while the remaining training images are treated as unlabeled to simulate real-world limited-label scenarios.

\textbf{Content Assessment Criteria.} Through discussions with anesthesiologists, gastric content assessment is formulated as a three-class classification task: Class I ($V \leq 50$ mL), Class II ($50 < V \leq 100$ mL), and Class III ($V > 100$ mL), where $V$ denotes the reference gastric content volume. In clinical practice, ground truth volumes are obtained by having fasting patients ingest a predefined amount of water. The distribution of samples across classes is summarized in Table~\ref{table:data_statistics}.

\subsection{Implementation Details}
The two stages of REASON are trained independently. All experiments are conducted using PyTorch 2.4.1 with Python 3.8 on Ubuntu 20.04.3. Models are trained on a single NVIDIA GeForce RTX 4090 GPU with 24 GB memory. All ultrasound images are resized to $256 \times 256$ pixels before training.

\textbf{Semi-Supervised Segmentation Model.} Under the semi-supervised training paradigm, only annotations for 10\% of the training images are used, while the remaining images are unlabeled. The segmentation model is optimized for 30,000 iterations using stochastic gradient descent (SGD) with a momentum of 0.9, weight decay of $1 \times 10^{-4}$, and an initial learning rate of 0.01. The batch size is set to 24, including 12 labeled and 12 unlabeled images in each iteration.

\textbf{Dual-Branch Fusion Classifier.} The dual-branch fusion classifier is trained for 120 epochs using SGD with a learning rate of 0.01, momentum of 0.9, and weight decay of $1 \times 10^{-4}$. By default, the hyper-parameters $\gamma$ and $\beta$ are set to 0.5 and 0.7, respectively.

\textbf{Evaluation Metrics.} To comprehensively evaluate performance on gastric content assessment, four metrics are reported: accuracy (Acc.), precision (Pre.), recall (Rec.), and F1 score (F1).
\begin{align}
    &{\rm Acc.}=\frac{\rm TP+TN}{\rm TP+TN+FP+FN} \\
    &{\rm Pre.}=\frac{\rm TP}{\rm TP+FP} \\
    &{\rm Rec.}=\frac{\rm TP}{\rm TP+FN} \\
    &{\rm F1}=\frac{\rm 2 TP}{\rm2 TP+FP+FN}
\end{align}
where ${\rm TP}$, ${\rm TN}$, ${\rm FP}$, and ${\rm FN}$ denote true positives, true negatives, false positives, and false negatives, respectively.

Moreover, Dice Similarity Coefficient (DSC) is adopted to evaluate the performance of the segmentation model, defined as:
\begin{align}
    {\rm DSC}=2\times\frac{|P\cap G|}{|P|+|G|}
\end{align}
where $P$ and $G$ indicate the predicted mask and ground truth mask, respectively.

\subsection{Main Results}
The framework is compared with several state-of-the-art methods on the gastric content assessment task. For transformer-based models, the standard image size of $224 \times 224$ is used as per their typical architecture. Five-fold cross-validation is performed, and results are reported as ``${\rm mean}\pm{\rm std}$''. Notably, all baseline methods operate on a single view. For fairness, their average performance across the two image views (SUP and RLD) is reported, as summarized in Table~\ref{table:sota}.

\begin{figure*}[t]
\centering
\centerline{\includegraphics{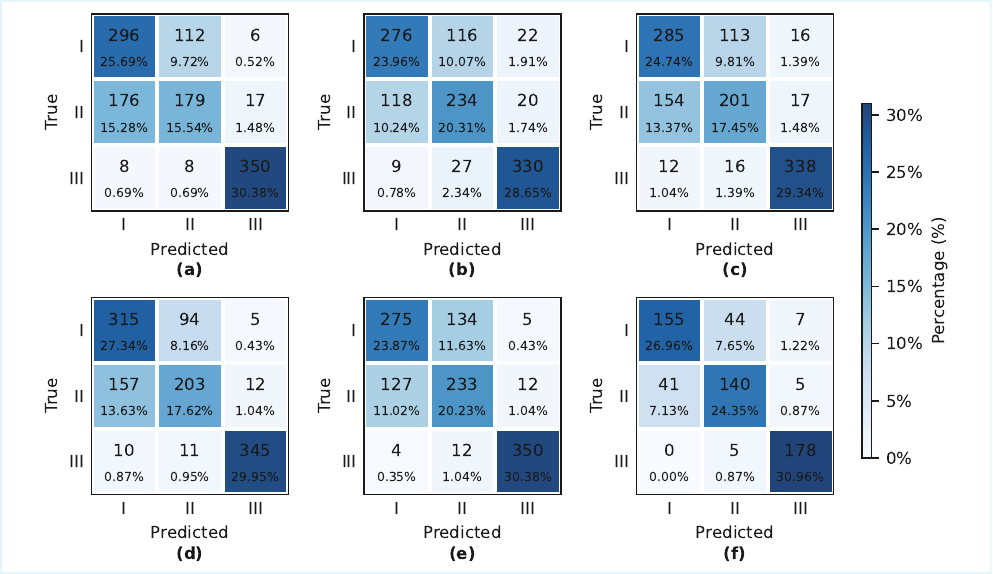}}
\caption{\textbf{Confusion matrices of several representative methods.} (a) DenseNet121. (b) EfficientNet-B0. (c) MobileNet-V3. (d) SoftAug. (e) MedMamba. \textbf{(f) REASON (Ours)}. Single view-based methods (a)-(e) sum ten confusion matrices (5 folds $\times$ 2 views), while our dual view-based REASON (f) sums five confusion matrices (5 folds).}
\label{fig:confusion_matrices}
\end{figure*}
\begin{figure*}[t]
\centering
\centerline{\includegraphics{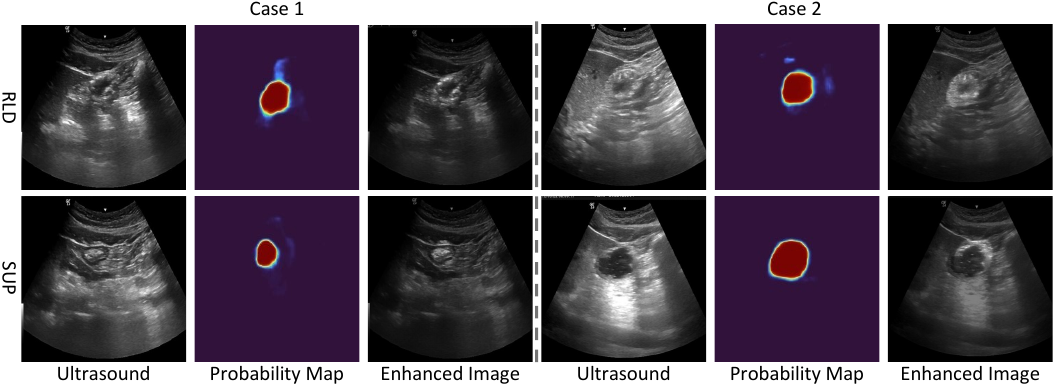}}
\caption{\textbf{Qualitative results of PMG.} The left block shows \textit{Case 1} and the right block shows \textit{Case 2}. Each case occupies two rows (RLD on top, SUP on bottom) and three columns: raw ultrasound (left), probability map (middle), and PMG-enhanced image (right). PMG highlights gastric regions while suppressing background artifacts.}
\label{fig:ablation_probability_map_guidance}
\end{figure*}
\begin{figure*}[t]
\centering
\centerline{\includegraphics{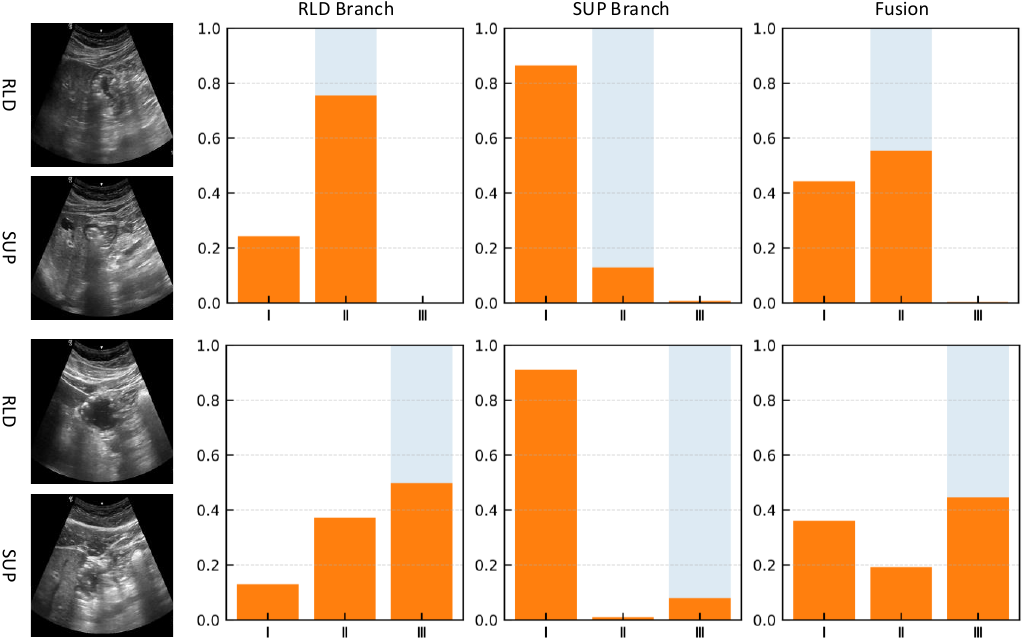}}
\caption{\textbf{Qualitative results of DBFC.} Bar chats represent the probability distributions from the RLD branch, the SUP branch, and DBFC. Ground truth classes are highlighted in blue.}
\label{fig:ablation_dual_branch_fusion}
\end{figure*}

\begin{table*}[t]
\caption{\textbf{Ablations on the proposed PMG and DBFC.} The best results are highlighted in \textbf{bold} and the second best results are \underline{underlined}.}
\label{table:ablation}
\centering
\renewcommand\arraystretch{1.2}
\begin{tabular}{llllll}
\toprule
PMG & DBFC & Acc. (\%) $\uparrow$ & Pre. (\%) $\uparrow$ & Rec. (\%) $\uparrow$ & F1 (\%) $\uparrow$ \\ \midrule
 & & 71.63{\tiny$\pm$2.86} & 73.03{\tiny$\pm$4.26} & 72.15{\tiny$\pm$3.30} & 71.35{\tiny$\pm$2.41} \\
 \checkmark & & 75.10{\tiny$\pm$1.96} {\color{blue}($\uparrow$3.47)} & 76.28{\tiny$\pm$1.57} {\color{blue}($\uparrow$3.25)} & 75.72{\tiny$\pm$1.98} {\color{blue}($\uparrow$3.57)} & 75.21{\tiny$\pm$1.59} {\color{blue}($\uparrow$3.86)} \\
 & \checkmark & \underline{77.12}{\tiny$\pm$4.90} {\color{blue}($\uparrow$5.49)} & \underline{77.84}{\tiny$\pm$4.48} {\color{blue}($\uparrow$4.81)} & \underline{77.78}{\tiny$\pm$4.64} {\color{blue}($\uparrow$5.63)} & \underline{77.34}{\tiny$\pm$4.49} {\color{blue}($\uparrow$5.99)} \\
\rowcolor{orange!20}\checkmark & \checkmark & \textbf{82.15}{\tiny$\pm$3.98} {\color{blue}($\uparrow$10.52)} & \textbf{83.23}{\tiny$\pm$2.89} {\color{blue}($\uparrow$10.20)} & \textbf{82.82}{\tiny$\pm$3.35} {\color{blue}($\uparrow$10.67)} & \textbf{82.10}{\tiny$\pm$3.98} {\color{blue}($\uparrow$10.75)} \\ 
\bottomrule
\end{tabular}
\end{table*}

Several key observations can be drawn from Table~\ref{table:sota}: \textbf{\textit{i)}} Learning-based methods outperform the empirical formula-based approach. By extracting high-level visual features from ultrasound, learning-based methods demonstrate greater robustness to variations in image quality and gastric deformation. \textbf{\textit{ii)}} Transformer-based methods perform poorly on this task. Both transformer-based models achieve accuracy close to random chance (33.3\%), even lower than the empirical formula-based method. Because gastric regions occupy only a small portion of the images, these models may fail to capture the fine-grained structural details required for reliable classification. In addition, transformer-based models are typically data-intensive and require substantially larger datasets than CNN-based methods to achieve stable performance~\cite{han2022survey}. \textbf{\textit{iii)}} The proposed REASON achieves the best performance. Compared with the empirical formula-based method, REASON yields improvements of +29.40\% in Acc., +30.75\% in Pre., +30.20\% in Rec., and +29.57\% in F1, highlighting its strong generalization ability. REASON also consistently outperforms state-of-the-art classification baselines. These gains are attributed to the two-stage design: probability map guidance helps the model focus more effectively on gastric regions, while the dual-branch fusion classifier effectively integrates complementary spatial and contextual information from the SUP and RLD views.

The confusion matrices of representative methods are shown in Fig.~\ref{fig:confusion_matrices}. Across all methods, Class III is predicted most accurately, whereas most errors arise from confusion between Classes I and II. This pattern is consistent with the criteria defined in Sec.~\ref{sec:criteria}, where Classes I and II differ only by small volume thresholds, leading to limited inter-class separability. Compared with baselines (a)-(e), REASON (f) substantially improves the recall for Class II, from 48.12-62.90\% to 75.27\%. It also markedly reduces cross-confusion between Classes I and II, highlighting the benefits of probability map guidance and dual-view fusion.

\textbf{Statistical Analysis.} To assess the significance of performance differences in Table~\ref{table:sota}, paired t-tests are conducted comparing REASON with the best-performing baseline. The results show $p$-values < 0.01 for Acc., Pre., and F1, indicating statistically significant improvements. For Rec., the $p$-value was 0.079, slightly above the conventional threshold of 0.05. Considering that three out of four metrics achieved strong significance, it can be concluded that REASON outperforms the best baseline overall with statistical confidence.

\subsection{Ablation Study}
In this section, extensive experiments are conducted to evaluate the effectiveness of the proposed PMG and DBFC. Default settings are highlighted in \sethlcolor{orange!20}\hl{orange}. Five-fold cross-validation is performed, and results are reported as ``${\rm mean}\pm{\rm std}$''. For the model \textit{w/o} DBFC, the average performance across the two image views is reported. The quantitative results are summarized in Table~\ref{table:ablation}.

\begin{table*}[t]
\caption{\textbf{Ablations on training settings of the segmentation model.} U-Net~\cite{ronneberger2015u} is a fully-supervised setting. The best results are highlighted in \textbf{bold} and the second best results are \underline{underlined}.}
\label{table:seg_model}
\centering
\renewcommand\arraystretch{1.2}
\begin{tabular}{llllllll}
\toprule
Method & Labeled & Unlabeled & DSC (\%) $\uparrow$ & Acc. (\%) $\uparrow$ & Pre. (\%) $\uparrow$ & Rec. (\%) $\uparrow$ & F1 (\%) $\uparrow$ \\ 
\midrule
U-Net \cite{ronneberger2015u} {\tiny \color{gray}[MICCAI'15]} & 100\% & 0\% & \textbf{87.06} & 77.99{\tiny$\pm$5.78} & 79.62{\tiny$\pm$4.62} & 78.52{\tiny$\pm$5.79} & 77.94{\tiny$\pm$5.86} \\ 
U-Net \cite{ronneberger2015u} {\tiny \color{gray}[MICCAI'15]} & 10\% & 0\% & 77.81 & 78.30{\tiny$\pm$2.56} & 79.23{\tiny$\pm$2.82} & 78.81{\tiny$\pm$2.03} & 78.06{\tiny$\pm$3.09} \\ 
\midrule
SCP \cite{zhang2023self} {\tiny \color{gray}[MICCAI'23]} & 10\% & 90\% & 63.03 & \underline{78.48{\tiny$\pm$1.20}} & \underline{79.53{\tiny$\pm$1.91}} & \underline{78.83{\tiny$\pm$0.85}} & \underline{78.37{\tiny$\pm$1.33}} \\ 
FFT \cite{hu2025beta} {\tiny \color{gray}[CVPR'25]} & 10\% & 90\% & 81.82 & 73.45{\tiny$\pm$1.90} & 74.55{\tiny$\pm$1.51} & 73.99{\tiny$\pm$2.04} & 73.59{\tiny$\pm$1.25} \\ 
\rowcolor{orange!20}
BCP \cite{bai2023bidirectional} {\tiny \color{gray}[CVPR'23]} & 10\% & 90\% & \underline{82.98} & \textbf{82.15}{\tiny$\pm$3.98} & \textbf{83.23}{\tiny$\pm$2.89} & \textbf{82.82}{\tiny$\pm$3.35} & \textbf{82.10}{\tiny$\pm$3.98} \\ 
\bottomrule
\end{tabular}
\end{table*}

\textbf{Effects of Probability Map Guidance.} Using images enhanced by probability maps as classifier inputs yields clear improvements, \textit{i.e.}, +3.47\% in Acc., +3.25\% in Pre., +3.57\% in Rec., and +3.86\% in F1. As illustrated in Fig.~\ref{fig:ablation_probability_map_guidance}, PMG enhances gastric regions and guides the model to focus on learning more robust representations.

\textbf{Role of the Dual-Branch Fusion Classifier.} Incorporating DBFC into the baseline also produces notable gains, \textit{i.e.}, +5.49\% in Acc., +4.81\% in Pre., +5.63\% in Rec., and +5.99\% in F1. Fig.~\ref{fig:ablation_dual_branch_fusion} shows two representative cases. Because of variability in gastric anatomy, the output of a single branch may not align well with the ground truth. By fusing logits from both branches, the model reduces misclassifications and achieves more stable outputs.

After integrating PMG and DBFC, the model outperforms all variants, demonstrating that the two modules provide complementary benefits.

\begin{figure*}[t]
\centering
\centerline{\includegraphics{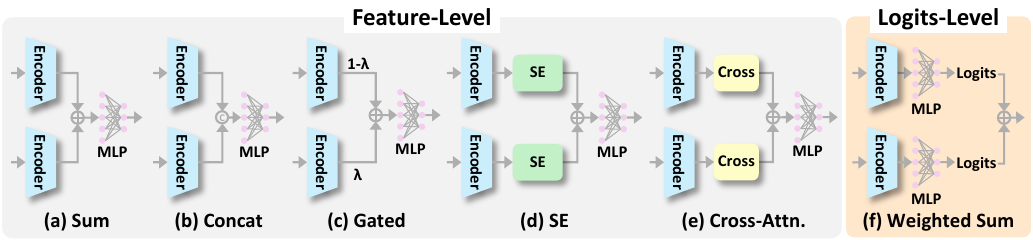}}
\caption{\textbf{Illustration of different fusion strategies.} A logits-level fusion strategy is compared with five feature-level alternatives. In (c), $\lambda$ is a learnable parameter. SE: Squeeze-and-Excitation; Cross-Attn.: Cross-Attention.}
\label{fig:fusion}
\end{figure*}
\begin{table*}[t]
\caption{\textbf{Ablations on classifier backbones.} The best results are highlighted in \textbf{bold} and the second best results are \underline{underlined}.}
\label{table:classifier}
\centering
\renewcommand\arraystretch{1.2}
\begin{tabular}{lllll}
\toprule
Backbone & Acc. (\%) $\uparrow$ & Pre. (\%) $\uparrow$ & Rec. (\%) $\uparrow$ & F1 (\%) $\uparrow$ \\ \midrule
VGG16 \cite{simonyan2014very} {\tiny \color{gray}[ICLR'15]} & 58.34{\tiny$\pm$8.42} & 58.18{\tiny$\pm$8.69} & 59.00{\tiny$\pm$8.37} & 57.93{\tiny$\pm$8.79} \\ 
ResNet50 \cite{he2016deep} {\tiny \color{gray}[CVPR'16]} & \underline{79.35{\tiny$\pm$3.01}} & \underline{80.57{\tiny$\pm$2.41}} & \underline{80.38{\tiny$\pm$2.77}} & \underline{79.47{\tiny$\pm$2.93}} \\
ResNet101 \cite{he2016deep} {\tiny \color{gray}[CVPR'16]} & 76.39{\tiny$\pm$3.35} & 77.69{\tiny$\pm$4.06} & 76.99{\tiny$\pm$3.51} & 76.50{\tiny$\pm$3.68} \\
EfficientNet-B0 \cite{tan2019efficientnet} {\tiny \color{gray}[ICML'19]} & 73.62{\tiny$\pm$5.43} & 74.96{\tiny$\pm$6.03} & 74.19{\tiny$\pm$5.12} & 73.41{\tiny$\pm$5.17} \\
EfficientNet-B5 \cite{tan2019efficientnet} {\tiny \color{gray}[ICML'19]} & 71.18{\tiny$\pm$2.39} & 73.07{\tiny$\pm$2.60} & 71.77{\tiny$\pm$2.60} & 71.31{\tiny$\pm$1.82} \\
MobileNet-V3 \cite{howard2019searching} {\tiny \color{gray}[ICCV'19]} & 69.25{\tiny$\pm$4.52} & 70.04{\tiny$\pm$6.53} & 69.86{\tiny$\pm$4.84} & 68.62{\tiny$\pm$5.02} \\
\rowcolor{orange!20}
DenseNet121 \cite{huang2017densely} {\tiny \color{gray}[CVPR'17]} & \textbf{82.15{\tiny$\pm$3.98}} & \textbf{83.23{\tiny$\pm$2.89}} & \textbf{82.82{\tiny$\pm$3.35}} & \textbf{82.10{\tiny$\pm$3.98}} \\ 
\bottomrule
\end{tabular}
\end{table*}
\begin{table*}[t]
\caption{\textbf{Ablations on fusion strategies.} The best results are highlighted in \textbf{bold} and the second best results are \underline{underlined}.}
\label{table:fusion}
\centering
\renewcommand\arraystretch{1.2}
\begin{tabular}{llcllll}
\toprule
\multicolumn{2}{l}{Strategy} & $\Delta$ Params. (M) & Acc. (\%) $\uparrow$ & Pre. (\%) $\uparrow$ & Rec. (\%) $\uparrow$ & F1 (\%) $\uparrow$ \\ \midrule
{\color{gray!60}\textit{w/o} Fusion} & {\color{gray!60}-} & {\color{gray!60}-} & {\color{gray!60}75.10{\tiny$\pm$1.96}} & {\color{gray!60}76.28{\tiny$\pm$1.57}} & {\color{gray!60}75.72{\tiny$\pm$1.98}} & {\color{gray!60}75.21{\tiny$\pm$1.59}} \\
\multirow{5}{*}{Feature-Level} & Concat & +2.10 & 79.17{\tiny$\pm$1.58} & 80.44{\tiny$\pm$1.28} & 79.39{\tiny$\pm$1.69} & 79.23{\tiny$\pm$1.45} \\
 & Sum & 0 & 79.52{\tiny$\pm$3.30} & 80.43{\tiny$\pm$3.23} & 80.03{\tiny$\pm$3.08} & 79.50{\tiny$\pm$3.10} \\
 & Gated & $\sim$0 & \underline{80.05}{\tiny$\pm$2.86} & \underline{81.68}{\tiny$\pm$2.13} & \underline{81.02}{\tiny$\pm$2.39} & \underline{80.31}{\tiny$\pm$2.45} \\
 & SE~\cite{hu2018squeeze} & +3.15 & 78.49{\tiny$\pm$3.33} & 79.45{\tiny$\pm$2.92} & 79.11{\tiny$\pm$3.08} & 78.74{\tiny$\pm$2.78} \\
 & Cross-Att.~\cite{lin2022cat} & +10.49 & 77.77{\tiny$\pm$3.45} & 79.02{\tiny$\pm$3.61} & 78.22{\tiny$\pm$3.52} & 77.78{\tiny$\pm$3.90} \\
\rowcolor{orange!20} Logits-Level & Weighted Sum & 0 & \textbf{82.15}{\tiny$\pm$3.98} & \textbf{83.23}{\tiny$\pm$2.89} & \textbf{82.82}{\tiny$\pm$3.35} & \textbf{82.10}{\tiny$\pm$3.98} \\ \bottomrule
\end{tabular}
\end{table*}


\subsection{In-Depth Analysis}
A detailed analysis of the proposed REASON is presented next. Default settings are highlighted in \sethlcolor{orange!20}\hl{orange}. Five-fold cross-validation is performed, and results are reported as ``${\rm mean}\pm{\rm std}$''.

\textbf{Different Training Settings of the Segmentation Model.} The impact of probability maps generated under different training settings of the segmentation model is systematically evaluated. Within the same semi-supervised protocol, BCP achieves the best overall performance across all metrics in Table~\ref{table:seg_model}. Additionally, U-Net is trained in a fully supervised manner using 10\% and 100\% of the labeled data. The results show that incorporating unlabeled data with an appropriate semi-supervised strategy markedly improves performance over the small-label baseline, \textit{i.e.}, U-Net (10\% labeled, 0\% unlabeled). Interestingly, although U-Net trained with 100\% labels achieves the highest DSC, BCP delivers substantially better classification results. This finding suggests that probability maps generated through the semi-supervised approach are more discriminative for classification, even when pixel-level overlap is slightly lower.

\textbf{Different Classifier Backbones.} To identify the most suitable classifier within the framework, several representative backbones are evaluated. As presented in Table~\ref{table:classifier}, DenseNet121 achieves the best performance across all metrics and exhibits notably low standard deviations, indicating both effectiveness and stability. Compared with the strongest competitor, ResNet50, DenseNet121 provides consistent gains of +2.80\% in Acc., +2.66\% in Pre., +2.44\% in Rec., and +2.63\% in F1.

\textbf{Feature-Level Fusion \textit{vs.} Logits-Level Fusion.} The proposed logits-level fusion strategy is compared with five carefully designed feature-level fusion approaches, as illustrated in Fig.~\ref{fig:fusion}. Quantitative results are summarized in Table~\ref{table:fusion}. Relative to the baseline \textit{w/o} fusion, both feature-level and logits-level fusion strategies yield clear performance gains, \textit{e.g.,} +2.67\%$\sim$+7.05\% Acc., underscoring the importance of integrating multi-view information. Moreover, the logits-level fusion strategy achieves the best overall performance while maintaining higher parameter efficiency than the feature-level alternatives.

\begin{figure*}[t]
\centering
\centerline{\includegraphics{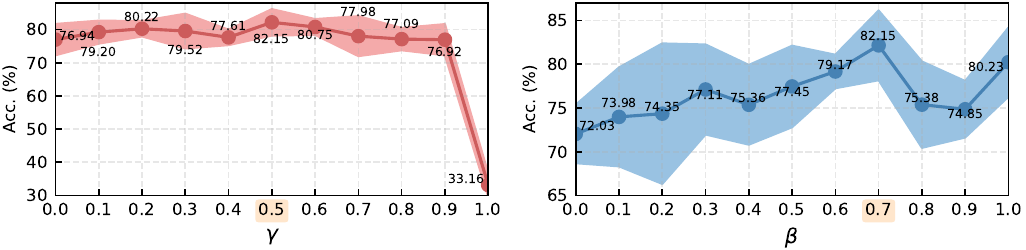}}
\caption{\textbf{Ablations on hyper-parameters $\gamma$ and $\beta$.}}
\label{fig:ablation_weight}
\end{figure*}

\begin{figure}[t]
\centering
\centerline{\includegraphics{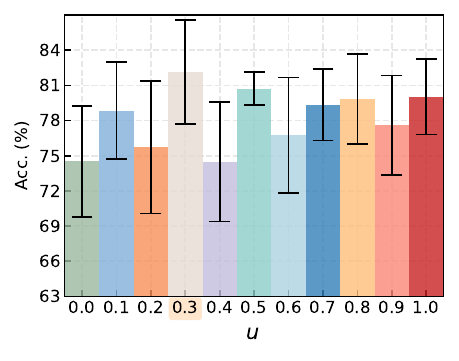}}
\caption{\textbf{Ablations on hyper-parameters $u$.}}
\label{fig:ablation_uweight}
\end{figure}

\textbf{Sensitivity to Hyper-Parameters.} In Fig.~\ref{fig:ablation_weight} and Fig.~\ref{fig:ablation_uweight}, the effects of the hyper-parameters $\gamma$, $\beta$, and $u$ on classification performance are analyzed. REASON remains robust for $\gamma \in [0,0.9]$, but performance drops markedly when $\gamma = 1$, where the DBFC inputs are defined as $x^\prime = x \odot p$. This indicates that preserving the original semantic content of ultrasound images is important for accurate classification. For the hyper-parameter $\beta$, REASON achieves the best performance when $\beta$ is relatively large, suggesting that the RLD view contributes more discriminative information than the SUP view. This finding is consistent with clinical observations~\cite{van2014ultrasound}.

For $u$, which balances the fused loss and the individual-branch losses, REASON exhibits stable performance within a moderate range. When $u=0$ ({\textit{i.e.}}, supervision is applied only to the fused branch), performance drops to $74.51\%$, indicating that supervision from the individual branches is essential for effective learning. Conversely, excessively large $u$ leads to instability, implying that over-emphasizing the individual-branch losses interferes with optimization of the fused prediction. Overall, assigning a moderate weight to individual-branch supervision enhances both performance and robustness.

\section{Conclusion}~\label{sec:conclusion}
This paper proposes REASON, a learning-based framework for gastric content assessment. The framework combines probability map guidance, which enhances focus on gastric-relevant regions, with a dual-branch fusion classifier that integrates information from SUP and RLD views to learn more discriminative representations. Experimental results demonstrate that REASON achieves state-of-the-art performance. This approach shows strong potential for clinical application, and future work will evaluate its generalizability across multi-center datasets and diverse patient populations.

\printcredits

\section*{Acknowledgments}
This work was supported in part by the National Key Research and Development Program of China under Grant 2023YFC2415100, in part by the National Natural Science Foundation of China under Grant 62373351, Grant 82327801, Grant 62073325, Grant 62303463, in part by the Chinese Academy of Sciences Project for Young Scientists in Basic Research under Grant No. YSBR-104, in part by the Beijing Natural Science Foundation under Grant F252068, Grant 4254107, in part by Beijing Nova Program under Grant 20250484813, in part by China Postdoctoral Science Foundation under Grant 2024M763535, in part by the Postdoctoral Fellowship Program of CPSF under Grant GZC20251170.

\bibliographystyle{model1-num-names}

\bibliography{cas-refs}

\end{document}